\documentclass[a4paper]{article}

\usepackage{INTERSPEECH2018}
\usepackage{url}

\title{ESPnet: End-to-End Speech Processing Toolkit}
\name{Shinji Watanabe$^1$, Takaaki Hori$^2$, Shigeki Karita$^3$, Tomoki Hayashi$^4$, Jiro Nishitoba$^5$, Yuya Unno$^6$, Nelson Enrique Yalta Soplin$^7$, Jahn Heymann$^8$, Matthew Wiesner$^1$, Nanxin Chen$^1$, Adithya Renduchintala$^1$, Tsubasa Ochiai$^9$, }
\address{
  $^1$Johns Hopkins University, 
  $^2$Mitsubishi Electric Research Laboratories,
  $^3$NTT Communication Science Laboratories, 
  $^4$Nagoya University, 
  $^5$Retrieva, Inc., 
  $^6$Preferred Networks, Inc.,
  $^7$Waseda University,
  $^8$Paderborn University,
  $^9$Doshisha University
  }
\email{shinjiw@jhu.edu}

\begin{document}

\maketitle
\begin{abstract}
This paper introduces a new open source platform for end-to-end speech processing named ESPnet.
ESPnet mainly focuses on end-to-end automatic speech recognition (ASR), and adopts widely-used dynamic neural network toolkits, Chainer and PyTorch, as a main deep learning engine.
ESPnet also follows the Kaldi ASR toolkit style for data processing, feature extraction/format, and recipes to provide a complete setup for speech recognition and other speech processing experiments.
This paper explains a major architecture of this software platform, several important functionalities, which differentiate ESPnet from other open source ASR toolkits, and experimental results with major ASR benchmarks.
\end{abstract}

\section{Introduction}
Automatic speech recognition (ASR) becomes a mature technology with a lot of research and development efforts mainly in speech processing communities.
Especially, these efforts have been driven by popular products including Google voice search, Amazon Alexa, and Apple Siri and open source activities including Kaldi \cite{Povey_ASRU2011}, HTK \cite{young2002htk}, Sphinx \cite{lee1990overview}, Julius \cite{lee2001julius}, RASR \cite{rybach2009rwth} in addition to general research activities.
These open source toolkits include feature extraction, acoustic modeling based on a hidden Markov model (HMM), Gaussian mixture model, and deep neural network (DNN), and decoding\footnote{Language modeling is often performed by external language model toolkits, for example SRILM \cite{stolcke2002srilm}}, and these enable us to use a full set of state-of-the-art ASR research and development achievement.

This paper describes a new open source toolkit named ESPnet (End-to-end speech processing toolkit), which aims to provide a neural end-to-end platform for ASR and other speech processing.
Unlike the above open source tools based on hybrid DNN/HMM architecutres \cite{hinton2012deep}, ESPnet provides a single neural network architecture to perform speech recognition in an end-to-end manner.
ESPnet adopts widely-used dynamic neural network toolkits, Chainer \cite{tokui2015chainer} and PyTorch \cite{paszke2017automatic}, as a main deep learning engine.
ESPnet also follows the style of Kaldi ASR toolkit \cite{Povey_ASRU2011} for data processing, feature extraction/format, and recipes to provide a complete setup for speech recognition and other speech processing experiments.

ESPnet fully utilizes benefits of two major end-to-end ASR implementations based on both connectionist temporal classification (CTC) \cite{graves2014towards,miao2015eesen,amodei2015deep} and attention-based encoder-decoder network \cite{chorowski2014end,lu2016training,chan2015listen,chiu2017state}.
Attention-based methods use an attention mechanism to perform alignment between acoustic frames and recognized symbols, while CTC uses Markov assumptions to efficiently solve sequential problems by dynamic programming. 
ESPnet adopts hybrid CTC/attention end-to-end ASR \cite{watanabe2017hybrid}, which effectively utilizes the advantages of both architectures in training and decoding. 
During training, we employ the multiobjective learning framework to improve robustness on irregular alignments and achieve fast convergence. 
During decoding, we perform joint decoding by combining both attention-based and CTC scores in a one-pass beam search algorithm to further eliminate irregular alignments.

In addition to the above basic architecture, ESPnet supports a number of end-to-end ASR techniques including a fusion of recurrent neural network language model (RNNLM) \cite{watanabe2017hybrid}, fast CTC computation by using the warp CTC library \cite{amodei2015deep}, many variations of attention methods.
With these state-of-the-art end-to-end ASR techniques, ESPnet also provides a number of recipes for major ASR benchmarks including Wall Street Journal (WSJ) \cite{paul1992design}, Librispeech \cite{panayotov2015librispeech}, TED-LIUM \cite{rousseau2012ted}, Corpus of Spontaneous Japanese (CSJ) \cite{maekawa2000spontaneous}, AMI \cite{hain2007ami}, HKUST Mandarin CTS \cite{liu2006hkust}, VoxForge \cite{voxforge}, CHiME-4/5 \cite{vincent2017analysis,barker2018fifth}, etc.
Thus, ESPnet provides publicly available state-of-the-art end-to-end ASR setups, which aim to accelerate the development of this emergent field.
This paper describes its basic architecture, functionalities, and benchmark results.
Note that several benchmarks including HKUST and CSJ score comparable/superior performance to the state-of-the-art hybrid DNN/HMM systems based on lattice-free maximum mutual information training \cite{povey2016purely}.
\section{Related studies}
This section mainly focuses on the comparison of ESPnet with publicly available toolkits within an \textit{end-to-end} ASR framework.
We can categorize the toolkits into two types based on CTC and attention architectures as follows:
\begin{itemize}
\item CTC-based: \\
EESEN \cite{miao2015eesen}, Stanford CTC \cite{lexfree2015}, Baidu Deepsppech \cite{amodei2015deep}, 
\item Attention-based: \\
Attention-LVCSR \cite{7472618}, OpenNMT speech to text \cite{klein2017opennmt}
\end{itemize}
Note that most of end-to-end ASR toolkits are based on CTC, while ESPnet is based on an attention-based encoder-decoder network.
Compared with Attention-LVCSR and OpenNMT, ESPnet has more specific functions to ASR applications including hybrid CTC/attention to deal with monotonic attentions, use of RNNLM during decoding, and a number of Kaldi-style ASR recipes, which make ESPnet unique to the other toolkits. 
\section{Functionality}
\begin{figure}[t]
  \centering
  \includegraphics[width=\linewidth]{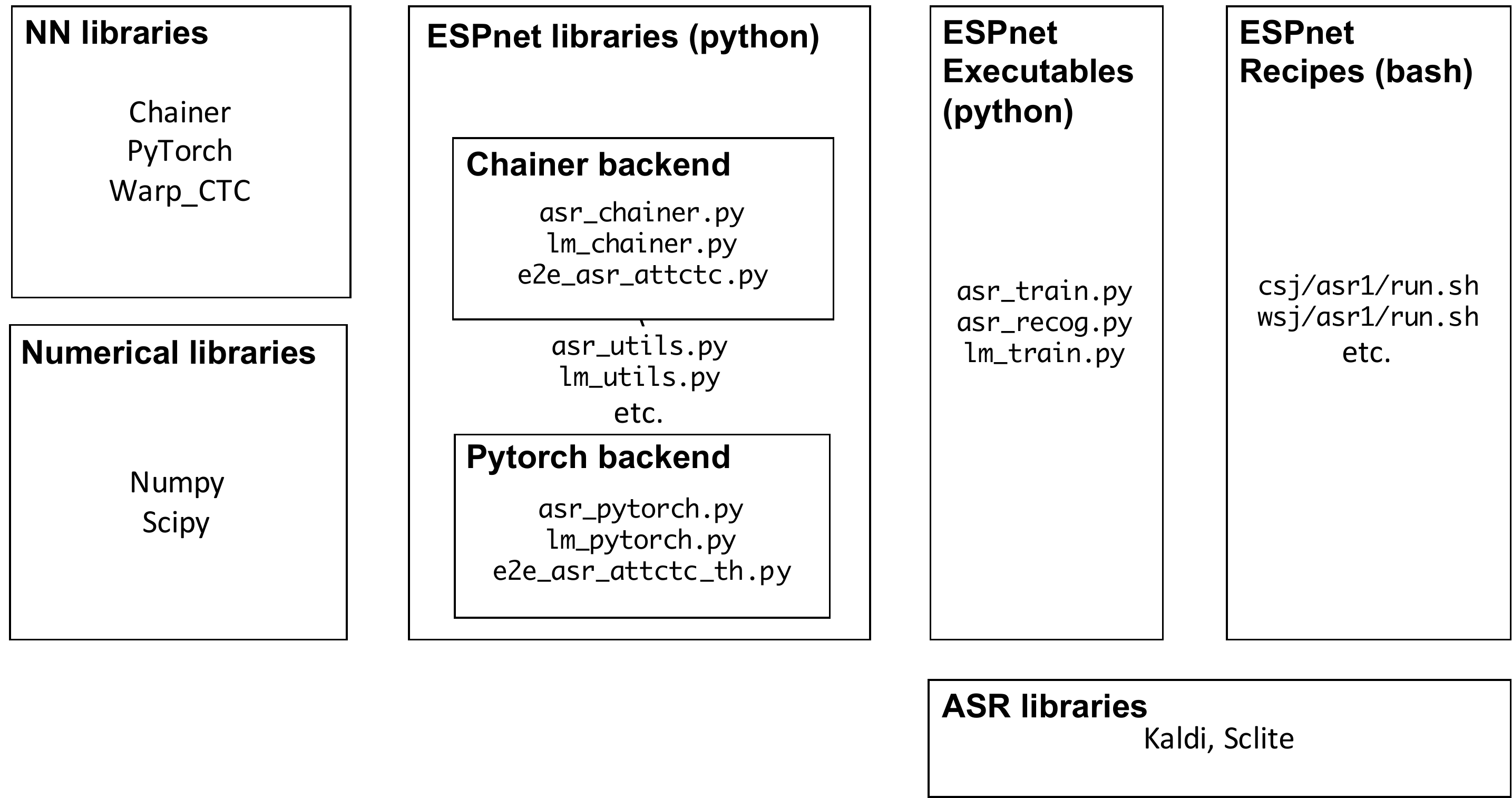}
  \caption{Software architecture of ESPnet.}
  \label{fig:system}
\end{figure}
Figure \ref{fig:system} shows a software architecture of ESPnet.
In the ESPnet, main neural network training and recognition parts are written in python, which calls Chainer and PyTorch by switching the backend option.
We also provide complete recipes to perform ASR experiments, which are written in the bash scripts by following the Kaldi manner.
The following sections describe several unique functions of ESPnet from existing other toolkits.
\subsection{Kaldi style data preprocessing}
ESPnet tightly integrates its data preprocessing part with Kaldi so that 1) we can fairly compare the performance obtained by Kaldi hybrid systems with ESPnet end-to-end systems and 2) we can make use of data preprocessing developed in the Kaldi recipe.
ESPnet also uses Kaldi feature extraction for most of recipes, although multichannel end-to-end ASR \cite{pmlr-v70-ochiai17a} includes speech enhancement and feature extraction with its network.

\subsection{Attention-based encoder-decoder}

\subsubsection{Encoder}
The default encoder network is represented by bidirectional long short-term memory (BLSTM) with subsampling (called pyramid BLSTM \cite{chan2015listen}) given $T$-length speech feature sequence $\mathbf{o} _{1:T}$ to extract high-level feature sequence $\mathbf{h} _{1:T'}$ as
\begin{equation}
 \mathbf{h} _{1:T'} = \text{BLSTM} (\mathbf{o} _{1:T}),
\end{equation}
where $T' < T$ in general due to the subsampling.
The Chainer backend also supports convolutional neural networks based on initial two blocks of VGG layer ($\text{VGG}_2 ()$) \cite{simonyan2014very} followed by BLSTM layers inspired by \cite{zhang2017very,hori2017advances}, that is
\begin{equation}
 \mathbf{h} _{1:T'} = \text{BLSTM} (\text{VGG}_2 (\mathbf{o} _{1:T})).
\end{equation}
This yields better performance than the pyramid BLSTM in many cases.

\subsubsection{Attention}
ESPnet uses a location-aware attention mechanism \cite{chorowski2015attention}, as a default attention.
A dot-product attention~\cite{luong2015effective} is also supported.
While the location-aware attention yields better performance, the dot-product attention is much faster in terms of the computational cost.
In addition to above attentions, the PyTorch backend supports more than 11 types of attention functions including additive attention~\cite{bahdanau2014neural}, coverage mechanism~\cite{see2017get}, and multi-head attention~\cite{vaswani2017attention}.

\subsection{Hybrid CTC/attention}
ESPnet adopts hybrid CTC/attention end-to-end ASR \cite{watanabe2017hybrid}, which effectively utilizes the advantages of both architectures in training and decoding. 

\subsubsection{Multiobjective training}
During training, we employ the multi objective learning framework by combining CTC $\mathcal{L} ^{\text{ctc}}$ and attention-based cross entropy $\mathcal{L} ^{\text{att}}$ to improve robustness and achieve fast convergence, as follows:
\begin{equation}
\mathcal{L} = \alpha \mathcal{L} ^{\text{ctc}} + (1 - \alpha) \mathcal{L} ^{\text{att}}
\end{equation}
This training method shares the same encoder with CTC and attention decoder networks. 
We have one tuning parameter $\alpha$ to linearly interpolate both objective functions and usually set as $\alpha = 0.5$ (equal contributions).

To alleviate overfitting problems, label smoothing techniques are available during training, which smooth the target distribution by dividing the probability mass for the correct label and the remaining labels in a certain ratio. We implemented unigram smoothing, where the distribution of remaining labels is set to be proportional to the unigram distribution of the labels \cite{pereyra2017regularizing}.

\subsubsection{Warp CTC}
CTC is one of the dominant parts for whole computation time in the training.
We use a warp CTC library developed by \cite{amodei2015deep} for both Chainer and PyTorch backends, which yields 5-10\% speed improvement in the total training time, compared with build-in CTC in the Chainer backend case.
\subsubsection{Joint decoding}
During decoding, we perform joint decoding by combining both attention-based and CTC scores in a one-pass beam search algorithm to further eliminate irregular alignments.
Let $y_n$ be a hypothesis of output label at position $n$ given a history $y_{1:n-1}$ and encoder output $\mathbf{h} _{1:T'}$.
The following score combination with attention $p^{\text{att}}$ and CTC  $p^{\text{ctc}}$ log probabilities is performed during the beam search:
\begin{align}
 & \log p^{\text{hyb}}(y_n|y_{1:n-1}, \mathbf{h} _{1:T'}) \nonumber \\
 & \quad = \alpha \log p ^{\text{ctc}} (y_n|y_{1:n-1}, \mathbf{h} _{1:T'}) \nonumber \\
 & \quad \quad + (1 - \alpha) \log p^{\text{att}}(y_n|y_{1:n-1}, \mathbf{h} _{1:T'}).
\end{align}
This hybrid CTC/attention architecture (multiobjective learning during training and joint decoding during recognition) is proposed in \cite{watanabe2017hybrid}, and a unique function compared with the other end-to-end ASR systems.
\subsection{Use of language model}
One of the most demanded functions of attention-based end-to-end ASR is how to make use of a language model trained with large amount of text corpora.
ESPnet can combine the log probability $p^{\text{lm}}$ of RNNLM during decoding as follows: 
\begin{align}
 & \log p(y_n|y_{1:n-1}, \mathbf{h} _{1:T'}) \nonumber \\
 & \quad = \log p ^{\text{hyb}} (y_n|y_{1:n-1}, \mathbf{h} _{1:T'}) + \beta \log p^{\text{lm}}(y_n|y_{1:n-1}).
\end{align}
$\beta$ is an additional scaling parameter.
This method corresponds to a shallow fusion of a decoder network and RNNLM originally proposed in neural machine translation \cite{gulcehre+al-2016-monolingual} and applied to end-to-end speech recognition \cite{hori2017advances}.

\subsection{ASR setup in adverse environments}
Although most of ASR recipes supported in ESPnet are standard English tasks, current ESPnet recipes deal with other languages including Japanese (CSJ), Mandarin Chinese (HKUST CTS), and other European languages through VoxForge.
With these various recipes, ESPnet can also realize multilingual end-to-end ASR system (e.g., 10 languages) by following our previous study \cite{watanabe2017language}.
In addition, the ESPnet recipes also include noise robust/far-field speech recognition tasks including AMI \cite{hain2007ami}, CHiME-4 \cite{vincent2017analysis}, and CHiME-5 tasks \cite{barker2018fifth}.
Especially ESPnet is an official end-to-end ASR baseline for the CHiME-5 challenge.

\section{Implementation}
\subsection{Standard recipe flow}
\begin{figure}[t]
  \centering
  \includegraphics[width=\linewidth]{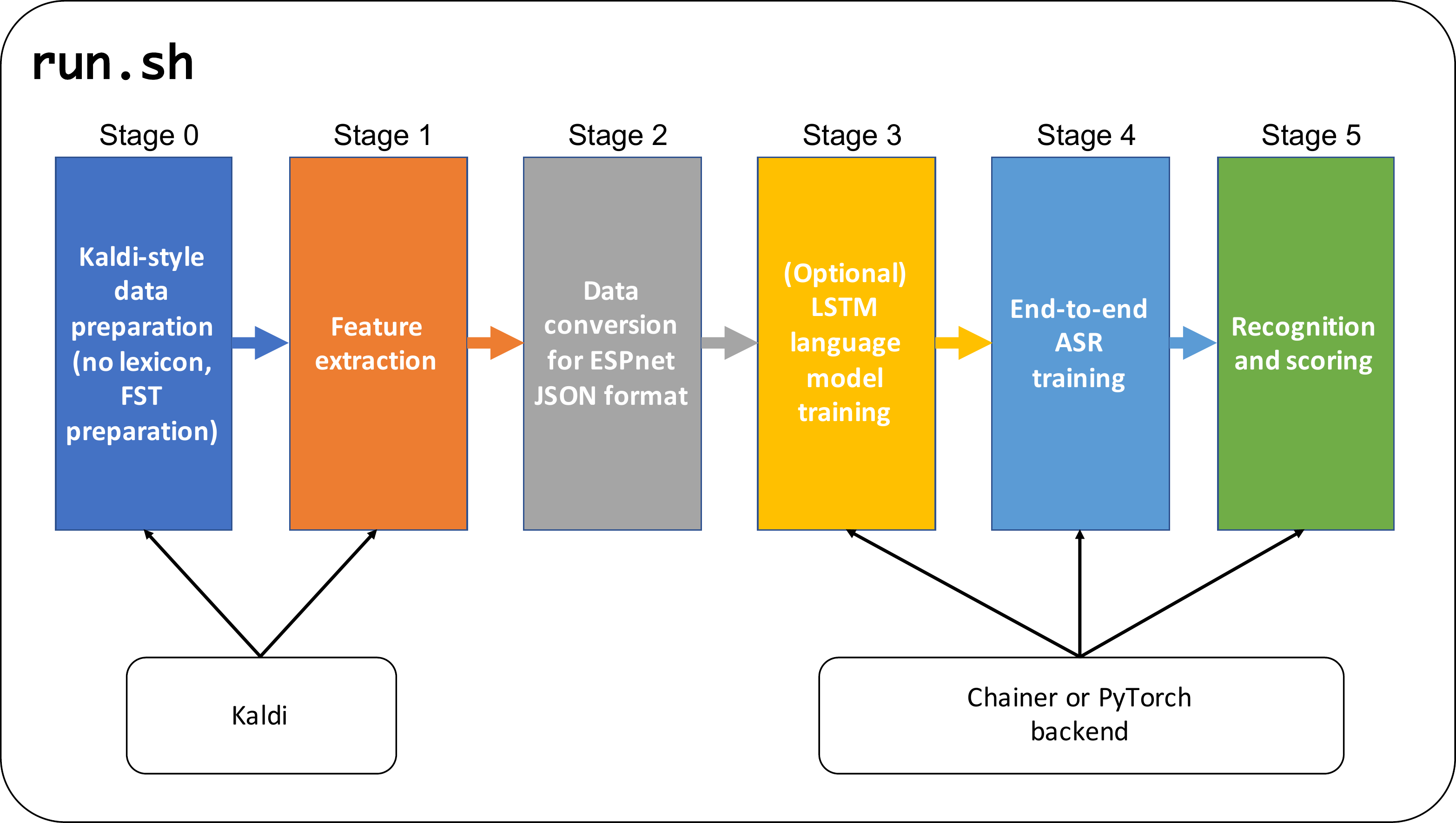}
  \caption{Experimental flow of standard ESPnet recipe.}
  \label{fig:recipe}
\end{figure}

Figure \ref{fig:recipe} shows a flow of standard recipes in ESPnet. 
The recipe is significantly simplified thanks to the benefit of end-to-end ASR, e.g., it does not have to include lexicon preparation, finite state transducer (FST) compilation, training/alignment based on HMM and Gaussian mixture modeling, and lattice generation for sequence discriminative training.

The standard recipe includes the following 6 stages in \verb|run.sh|\footnote{Several recipes including AMI, Librispeech, TED-LIUM, and VoxForge have an additional data downloading stage (\textbf{stage -1}).}:
\begin{itemize}
\item[\textbf{Stage 0}] Data preparation: 
We adopt the Kaldi data directory format, and we can simply use the Kaldi data preparation script (e.g., \verb|data_prep.sh|).
\item[\textbf{Stage 1}] Feature extraction: Again, we use the Kaldi feature extraction. 
Most of recipes use the 80-dimensional log Mel feature with the pitch feature (totally 83 dimensions).
\item[\textbf{Stage 2}] Data preparation for ESPnet: 
This stage converts all the information including in the Kaldi data directory (transcriptions, speaker and language IDs, and input and output lengths) to one JSON file (\verb|data.json|) except for input features.
\item[\textbf{Stage 3}] Language model training: Character-based RNNLM is trained by using either Chainer or PyTorch backend.
This is an optional stage, and several recipes do not have this stage).
\item[\textbf{Stage 4}] End-to-end ASR training: Hybrid CTC/attention-based encoder-decoder is trained by using either Chainer or PyTorch backend.
\item[\textbf{Stage 5}] Recognition: Speech recognition is performed by using RNNLM and end-to-end ASR model obtained by stages 3 and 4, respectively.
\end{itemize}

\subsection{Code lines}
In addition to the actual experimental stage, ESPnet also simplifies its coding lines.
Table \ref{tab:code_lines} compared the main source code of Kaldi, Julius, and ESPnet.
ESPnet can realize speech recognition including trainer and recognizer functions by only using 5K lines of python codes compared with Kaldi and Julius, thanks to the simplification of end-to-end ASR and use of Chainer or PyTorch for neural network backends and Kaldi for data preparation and feature extraction\footnote{Since Kaldi and Julius have various function including online real-time modes and Windows interfaces unlike ESPnet, we cannot directly compare them with the source code lines.}.
\begin{table}[tbh]
  \caption{Number of main source code lines of Kaldi, Julius, and ESPnet, and their main languages.}
  \label{tab:code_lines}
  \centering
  \begin{tabular}{c| c | c}
    \toprule
    Toolkit & \# lines & Language \\
   	\midrule
    Kaldi & 330K & c++ \\
    Julius & 60K & c \\
    ESPnet & 5.4K & python \\
    \bottomrule
  \end{tabular}
\end{table}

One of the most simplified module is a model representation part, since it does not have to explicitly represent a complicated speech recognition hierarchy from speech features, HMM states, context dependent phonemes, lexicons, to words.
This hierarchy is represented by a single neural network with at most thousand lines of python codes.
This also yields to simplify the recognition module with at most five hundred lines, as it is realized by a simple output-synchronous beam search.

\section{Experiments}
This section discusses the experimental results of our three main tasks, WSJ, CSJ, and HKUST.
The first experiment shows the effectiveness of the ESPnet with the famous WSJ tasks by using several experimental configurations, and also compare the  reports on the same task within an end-to-end ASR framework.
The other experiments compare the performance of ESPnet with state-of-the-art ASR systems for the CSJ and HKUST tasks.
The main reason for choosing these two languages is that these ideogram languages have relatively shorter lengths for letter sequences than those in alphabet languages, which greatly reduces the computational complexities, and makes it easy to handle context information in a decoder network. 
Actually, our prior investigation shows that Japanese and Mandarin Chinese end-to-end ASR can be easily scaled up, and shows reasonable performance without using various tricks developed for large-scale English tasks.

\begin{table}[tbh]
  \caption{Comparisons (CER, WER, and training time) of the WSJ task with other end-to-end ASR systems.}
  \label{tab:wsj}
  \centering
  \begin{tabular}{ l | c | c c }
    \toprule
    Method & Metric & dev93 & eval92 \\
   	\midrule
    ESPnet with VGG$_2$-BLSTM & CER & 10.1 & 7.6 \\
    + BLSTM layers (4 $\rightarrow$ 6) & CER & 8.5 & 5.9 \\
    + char-LSTMLM & CER & 8.3 & 5.2 \\
    + joint decoding & CER & 5.5 & 3.8 \\
    + label smoothing & CER & 5.3 & 3.6 \\
                     & WER & 12.4 & 8.9 \\
   	\midrule
    seq2seq + CNN (no LM) \cite{zhang2017very} & WER & N/A & 10.5 \\
    seq2seq + FST word LM \cite{chorowski2015attention} & CER & N/A & 3.9 \\
      & WER & N/A & 9.3 \\
    CTC + FST word LM \cite{miao2015eesen} & WER & N/A & 7.3 \\
    \bottomrule
  \end{tabular}
  \begin{tabular}{ l | c | c}
    \toprule
    Method & Wall Clock Time & \# GPUs \\
   	\midrule
    ESPnet (Chainer) & 20 hours & 1 \\
    ESPnet (PyTorch) & 5 hours & 1 \\
    seq2seq + CNN \cite{zhang2017very} & 120 hours & 10 \\
    \bottomrule
  \end{tabular}
\end{table}
Table \ref{tab:wsj} compares the performance of the ESPnet with different techniques in the WSJ task.
The use of a deeper encoder network, integration of character-based LSTMLM, and joint CTC/attention decoding steadily improved the performance.
Table \ref{tab:wsj} also compares the result of ESPnet with the other reports.
Since these reports are based on different conditions (e.g., \cite{zhang2017very} does not use any language models, while \cite{chorowski2015attention} and \cite{miao2015eesen} use a word-based language model through FST), we cannot directly compare them.
But we can state that ESPnet provides reasonable performance by comparing with these prior studies.
Table \ref{tab:wsj} also provides the computational time for main end-to-end ASR network training with number of GPUs.
ESPnet achieved very fast training especially for the PyTorch backend even with a single GPU (gtx1080ti), compared with \cite{zhang2017very} for the same WSJ task.

However, one of the issues of these end-to-end ASR systems is that their performance does not reach that of the state-of-the-art hybrid HMM/DNN systems.
For example, the WER of the hybrid HMM/DNN systems for the WSJ task is below 5\%, and this degradation probably comes from the lack of the amount of training data.
Actually, \cite{amodei2015deep} and \cite{chiu2017state} report comparable or superior performance to the state-of-the-art hybrid HMM/DNN systems in very large English tasks, although these results are not usually accomplished by many of research communities due to the lack of computational resources.
Therefore, scaling up the English task with keeping low computational resources or improving the performance by mitigating the data sparseness issue is one of our important future studies.

\begin{table}[tbh]
  \caption{Corpus of Spontaneous Japanese (CSJ) task (CER \%)}
  \label{tab:csj}
  \centering
  \begin{tabular}{ l | c c c }
    \toprule
     & eval1 & eval2 & eval3 \\
   	\midrule
     ESPnet & 8.7 & 6.2 & 6.9 \\
     ESPnet (5 GPUs) & 8.5 & 6.1 & 6.8 \\
     HMM/DNN (Kaldi nnet1) & 9.0 & 7.2 & 9.6 \\
     CTC-syllable \cite{kanda2016maximum} & 9.4 & 7.3 & 7.5 \\
     \bottomrule
  \end{tabular}
\end{table}

\begin{table}[tbh]
  \caption{HKUST Mandarin CTS task (CER \%).}
  \label{tab:hkust}
  \centering
  \begin{tabular}{ l | c}
    \toprule
     & eval \\
   	\midrule
    ESPnet & 28.3 \\
	HMM/LSTM (Kaldi nnet3) & 33.5 \\
	CTC with language model \cite{miao2015eesen} & 34.8 \\
	HMM/TDNN, LF MMI \cite{povey2016purely} & 28.2 \\
    \bottomrule
  \end{tabular}
\end{table}
Compared with the English tasks, end-to-end ASR systems can easily achieve comparable performance to the state-of-the-art hybrid HMM/DNN systems in the Japanese and Mandarin Chinese tasks.
Note that ESPnet does not use lexical information (pronunciation dictionary and morphological analyzer), which are essential components in the HMM/DNN and CTC-syllable systems.
Tables \ref{tab:csj} and \ref{tab:hkust} compare the best system of ESPnet (i.e., VGG$_2$-BLSTM, char-RNNLM, and joint decoding) with the hybrid HMM/DNN systems.
Especially, ESPnet almost reached the latest best performance of the HMM/DNN system with lattice-free MMI training \cite{povey2016purely} in the HKUST task.

\section{Conclusions}
This paper introduced a new end-to-end ASR toolkit named ESPnet.
ESPnet fully utilizes dynamic neural network toolkits, Chainer and PyTorch, as a main deep learning engine, and extremely simplifies training and recognition of the whole ASR pipeline.
A number of experiments and comparisons with other reports show that ESPnet achieves reasonable ASR performance and also reaches comparable performance to the state-of-the-art HMM/DNN systems with a legacy setup.
ESPnet has been actively developed, and multi-GPU function, data augmentation, multihead decoder, multichannel end-to-end ASR, and Babel multilingual ASR experiments are in preparation.
Especially with the multi-GPU function (5 GPUs), ESPnet finished the training of 581 hours of the CSJ task only with 26 hours.

\bibliographystyle{IEEEtran}

\bibliography{mybib}

\begin{thebibliography}{10}
\providecommand{\url}[1]{#1}
\csname url@samestyle\endcsname
\providecommand{\newblock}{\relax}
\providecommand{\bibinfo}[2]{#2}
\providecommand{\BIBentrySTDinterwordspacing}{\spaceskip=0pt\relax}
\providecommand{\BIBentryALTinterwordstretchfactor}{4}
\providecommand{\BIBentryALTinterwordspacing}{\spaceskip=\fontdimen2\font plus
\BIBentryALTinterwordstretchfactor\fontdimen3\font minus
  \fontdimen4\font\relax}
\providecommand{\BIBforeignlanguage}[2]{{%
\expandafter\ifx\csname l@#1\endcsname\relax
\typeout{** WARNING: IEEEtran.bst: No hyphenation pattern has been}%
\typeout{** loaded for the language `#1'. Using the pattern for}%
\typeout{** the default language instead.}%
\else
\language=\csname l@#1\endcsname
\fi
#2}}
\providecommand{\BIBdecl}{\relax}
\BIBdecl

\bibitem{Povey_ASRU2011}
D.~Povey, A.~Ghoshal, G.~Boulianne, L.~Burget, O.~Glembek, N.~Goel,
  M.~Hannemann, P.~Motlicek, Y.~Qian, P.~Schwarz, J.~Silovsky, G.~Stemmer, and
  K.~Vesely, ``The kaldi speech recognition toolkit,'' in \emph{IEEE Workshop
  on Automatic Speech Recognition and Understanding (ASRU)}, Dec. 2011.

\bibitem{young2002htk}
\BIBentryALTinterwordspacing
Y.~Young, G.~Evermann, D.~Kershaw, G.~Moore, J.~Odell, D.~Ollason, D.~Povey,
  V.~Valtchev, and P.~Woodland, ``The htk book,'' \emph{Cambridge university
  engineering department}, vol.~3, p. 175, 2002. [Online]. Available:
  \url{http://htk.eng.cam.ac.uk/}
\BIBentrySTDinterwordspacing

\bibitem{lee1990overview}
\BIBentryALTinterwordspacing
K.-F. Lee, H.-W. Hon, and R.~Reddy, ``An overview of the {SPHINX} speech
  recognition system,'' \emph{IEEE Transactions on Acoustics, Speech, and
  Signal Processing}, vol.~38, no.~1, pp. 35--45, 1990. [Online]. Available:
  \url{http://cmusphinx.sourceforge.net/}
\BIBentrySTDinterwordspacing

\bibitem{lee2001julius}
A.~Lee, T.~Kawahara, and K.~Shikano, ``Julius an open source real-time large
  vocabulary recognition engine,'' in \emph{Proc. Eurospeech}, 2001, pp.
  1691--1694.

\bibitem{rybach2009rwth}
\BIBentryALTinterwordspacing
D.~Rybach, C.~Gollan, G.~Heigold, B.~Hoffmeister, J.~L{\"o}{\"o}f,
  R.~Schl{\"u}ter, and H.~Ney, ``The {RWTH} aachen university open source
  speech recognition system.'' in \emph{Interspeech}, 2009, pp. 2111--2114.
  [Online]. Available: \url{https://www-i6.informatik.rwth-aachen.de/rwth-asr/}
\BIBentrySTDinterwordspacing

\bibitem{stolcke2002srilm}
\BIBentryALTinterwordspacing
A.~Stolcke \emph{et~al.}, ``{SRILM}-an extensible language modeling toolkit.''
  in \emph{Interspeech}, vol. 2002, 2002, pp. 901--904. [Online]. Available:
  \url{http://www.speech.sri.com/projects/srilm/}
\BIBentrySTDinterwordspacing

\bibitem{hinton2012deep}
G.~Hinton, L.~Deng, D.~Yu, G.~E. Dahl, A.-r. Mohamed, N.~Jaitly, A.~Senior,
  V.~Vanhoucke, P.~Nguyen, T.~N. Sainath \emph{et~al.}, ``Deep neural networks
  for acoustic modeling in speech recognition: The shared views of four
  research groups,'' \emph{IEEE Signal Processing Magazine}, vol.~29, no.~6,
  pp. 82--97, 2012.

\bibitem{tokui2015chainer}
S.~Tokui, K.~Oono, S.~Hido, and J.~Clayton, ``Chainer: a next-generation open
  source framework for deep learning,'' in \emph{Proceedings of workshop on
  machine learning systems (LearningSys) in the twenty-ninth annual conference
  on neural information processing systems (NIPS)}, vol.~5, 2015.

\bibitem{paszke2017automatic}
A.~Paszke, S.~Gross, S.~Chintala, G.~Chanan, E.~Yang, Z.~DeVito, Z.~Lin,
  A.~Desmaison, L.~Antiga, and A.~Lerer, ``Automatic differentiation in
  {PyTorch},'' in \emph{Proceedings of The future of gradient-based machine
  learning software and techniques (Autodiff) in the twenty-ninth annual
  conference on neural information processing systems (NIPS)}, 2017.

\bibitem{graves2014towards}
A.~Graves and N.~Jaitly, ``Towards end-to-end speech recognition with recurrent
  neural networks,'' in \emph{International Conference on Machine Learning
  (ICML)}, 2014, pp. 1764--1772.

\bibitem{miao2015eesen}
Y.~Miao, M.~Gowayyed, and F.~Metze, ``{EESEN}: End-to-end speech recognition
  using deep {RNN} models and {WFST}-based decoding,'' in \emph{IEEE Workshop
  on Automatic Speech Recognition and Understanding (ASRU)}, 2015, pp.
  167--174.

\bibitem{amodei2015deep}
\BIBentryALTinterwordspacing
D.~Amodei, R.~Anubhai, E.~Battenberg, C.~Case, J.~Casper, B.~Catanzaro,
  J.~Chen, M.~Chrzanowski, A.~Coates, G.~Diamos \emph{et~al.}, ``Deep speech 2:
  End-to-end speech recognition in english and mandarin,'' \emph{arXiv preprint
  arXiv:1512.02595}, 2015. [Online]. Available:
  \url{https://github.com/baidu-research/ba-dls-deepspeech}
\BIBentrySTDinterwordspacing

\bibitem{chorowski2014end}
J.~Chorowski, D.~Bahdanau, K.~Cho, and Y.~Bengio, ``End-to-end continuous
  speech recognition using attention-based recurrent {NN}: First results,''
  \emph{arXiv preprint arXiv:1412.1602}, 2014.

\bibitem{lu2016training}
L.~Lu, X.~Zhang, and S.~Renals, ``On training the recurrent neural network
  encoder-decoder for large vocabulary end-to-end speech recognition,'' in
  \emph{IEEE International Conference on Acoustics, Speech and Signal
  Processing (ICASSP)}, 2016, pp. 5060--5064.

\bibitem{chan2015listen}
W.~Chan, N.~Jaitly, Q.~V. Le, and O.~Vinyals, ``Listen, attend and spell: A
  neural network for large vocabulary conversational speech recognition,'' in
  \emph{IEEE International Conference on Acoustics, Speech and Signal
  Processing (ICASSP)}, 2015.

\bibitem{chiu2017state}
C.-C. Chiu, T.~N. Sainath, Y.~Wu, R.~Prabhavalkar, P.~Nguyen, Z.~Chen,
  A.~Kannan, R.~J. Weiss, K.~Rao, K.~Gonina \emph{et~al.}, ``State-of-the-art
  speech recognition with sequence-to-sequence models,'' \emph{arXiv preprint
  arXiv:1712.01769}, 2017.

\bibitem{watanabe2017hybrid}
S.~Watanabe, T.~Hori, S.~Kim, J.~R. Hershey, and T.~Hayashi, ``Hybrid
  {CTC}/attention architecture for end-to-end speech recognition,'' \emph{IEEE
  Journal of Selected Topics in Signal Processing}, vol.~11, no.~8, pp.
  1240--1253, 2017.

\bibitem{paul1992design}
D.~B. Paul and J.~M. Baker, ``The design for the {Wall} {Street}
  {Journal}-based {CSR} corpus,'' in \emph{Proceedings of the workshop on
  Speech and Natural Language}.\hskip 1em plus 0.5em minus 0.4em\relax
  Association for Computational Linguistics, 1992, pp. 357--362.

\bibitem{panayotov2015librispeech}
V.~Panayotov, G.~Chen, D.~Povey, and S.~Khudanpur, ``Librispeech: an {ASR}
  corpus based on public domain audio books,'' in \emph{IEEE International
  Conference on Acoustics, Speech and Signal Processing (ICASSP)}, 2015, pp.
  5206--5210.

\bibitem{rousseau2012ted}
A.~Rousseau, P.~Del{\'e}glise, and Y.~Esteve, ``{TED-LIUM}: an automatic speech
  recognition dedicated corpus.'' in \emph{International Conference on Language
  Resources and Evaluation (LREC)}, 2012, pp. 125--129.

\bibitem{maekawa2000spontaneous}
K.~Maekawa, H.~Koiso, S.~Furui, and H.~Isahara, ``Spontaneous speech corpus of
  {Japanese},'' in \emph{International Conference on Language Resources and
  Evaluation (LREC)}, vol.~2, 2000, pp. 947--952.

\bibitem{hain2007ami}
T.~Hain, L.~Burget, J.~Dines, G.~Garau, V.~Wan, M.~Karafi, J.~Vepa, and
  M.~Lincoln, ``The {AMI} system for the transcription of speech in meetings,''
  in \emph{IEEE International Conference on Acoustics, Speech and Signal
  Processing (ICASSP)}, 2007, pp. 357--360.

\bibitem{liu2006hkust}
Y.~Liu, P.~Fung, Y.~Yang, C.~Cieri, S.~Huang, and D.~Graff, ``{HKUST/MTS}: {A}
  very large scale {Mandarin} telephone speech corpus,'' in \emph{Chinese
  Spoken Language Processing}.\hskip 1em plus 0.5em minus 0.4em\relax Springer,
  2006, pp. 724--735.

\bibitem{voxforge}
``{VoxForge},'' http://www.voxforge.org/.

\bibitem{vincent2017analysis}
E.~Vincent, S.~Watanabe, A.~A. Nugraha, J.~Barker, and R.~Marxer, ``An analysis
  of environment, microphone and data simulation mismatches in robust speech
  recognition,'' \emph{Computer Speech \& Language}, vol.~46, pp. 535--557,
  2017.

\bibitem{barker2018fifth}
J.~Barker, S.~Watanabe, E.~Vincent, and J.~Trmal, ``The fifth {`CHiME’}
  speech separation and recognition challenge: Dataset, task and baselines,''
  in \emph{Interspeech}, 2018, (submitting).

\bibitem{povey2016purely}
D.~Povey, V.~Peddinti, D.~Galvez, P.~Ghahrmani, V.~Manohar, X.~Na, Y.~Wang, and
  S.~Khudanpur, ``Purely sequence-trained neural networks for asr based on
  lattice-free {MMI},'' in \emph{Interspeech}, 2016, pp. 2751--2755.

\bibitem{lexfree2015}
\BIBentryALTinterwordspacing
A.~L. Maas, Z.~Xie, D.~Jurafsky, and A.~Y. Ng, ``Lexicon-free conversational
  speech recognition with neural networks,'' in \emph{Proceedings the North
  American Chapter of the Association for Computational Linguistics (NAACL)},
  2015. [Online]. Available: \url{https://github.com/amaas/stanford-ctc}
\BIBentrySTDinterwordspacing

\bibitem{7472618}
\BIBentryALTinterwordspacing
D.~Bahdanau, J.~Chorowski, D.~Serdyuk, P.~Brakel, and Y.~Bengio, ``End-to-end
  attention-based large vocabulary speech recognition,'' in \emph{2016 IEEE
  International Conference on Acoustics, Speech and Signal Processing
  (ICASSP)}, March 2016, pp. 4945--4949. [Online]. Available:
  \url{https://github.com/rizar/attention-lvcsr}
\BIBentrySTDinterwordspacing

\bibitem{klein2017opennmt}
G.~Klein, Y.~Kim, Y.~Deng, J.~Senellart, and A.~M. Rush, ``Opennmt: Open-source
  toolkit for neural machine translation,'' \emph{arXiv preprint
  arXiv:1701.02810}, 2017.

\bibitem{pmlr-v70-ochiai17a}
T.~Ochiai, S.~Watanabe, T.~Hori, and J.~R. Hershey, ``Multichannel end-to-end
  speech recognition,'' in \emph{Proceedings of the 34th International
  Conference on Machine Learning (ICML)}, vol.~70, 2017, pp. 2632--2641.

\bibitem{simonyan2014very}
K.~Simonyan and A.~Zisserman, ``Very deep convolutional networks for
  large-scale image recognition,'' \emph{arXiv preprint arXiv:1409.1556}, 2014.

\bibitem{zhang2017very}
Y.~Zhang, W.~Chan, and N.~Jaitly, ``Very deep convolutional networks for
  end-to-end speech recognition,'' in \emph{IEEE International Conference on
  Acoustics, Speech and Signal Processing (ICASSP)}.\hskip 1em plus 0.5em minus
  0.4em\relax IEEE, 2017, pp. 4845--4849.

\bibitem{hori2017advances}
T.~Hori, S.~Watanabe, Y.~Zhang, and W.~Chan, ``Advances in joint
  {CTC}-attention based end-to-end speech recognition with a deep {CNN} encoder
  and {RNN-LM},'' in \emph{Interspeech}, 2017, pp. 949--953.

\bibitem{chorowski2015attention}
J.~K. Chorowski, D.~Bahdanau, D.~Serdyuk, K.~Cho, and Y.~Bengio,
  ``Attention-based models for speech recognition,'' in \emph{Advances in
  Neural Information Processing Systems (NIPS)}, 2015, pp. 577--585.

\bibitem{luong2015effective}
M.-T. Luong, H.~Pham, and C.~D. Manning, ``Effective approaches to
  attention-based neural machine translation,'' \emph{arXiv preprint
  arXiv:1508.04025}, 2015.

\bibitem{bahdanau2014neural}
D.~Bahdanau, K.~Cho, and Y.~Bengio, ``Neural machine translation by jointly
  learning to align and translate,'' \emph{arXiv preprint arXiv:1409.0473},
  2014.

\bibitem{see2017get}
A.~See, P.~J. Liu, and C.~D. Manning, ``Get to the point: Summarization with
  pointer-generator networks,'' \emph{arXiv preprint arXiv:1704.04368}, 2017.

\bibitem{vaswani2017attention}
A.~Vaswani, N.~Shazeer, N.~Parmar, J.~Uszkoreit, L.~Jones, A.~N. Gomez,
  {\L}.~Kaiser, and I.~Polosukhin, ``Attention is all you need,'' in
  \emph{Advances in Neural Information Processing Systems}, 2017, pp.
  6000--6010.

\bibitem{pereyra2017regularizing}
G.~Pereyra, G.~Tucker, J.~Chorowski, {\L}.~Kaiser, and G.~Hinton,
  ``Regularizing neural networks by penalizing confident output
  distributions,'' \emph{arXiv preprint arXiv:1701.06548}, 2017.

\bibitem{gulcehre+al-2016-monolingual}
{\c C}.~G{\"{u}}l{\c c}ehre, O.~Firat, K.~Xu, K.~Cho, L.~Barrault, H.-C. Lin,
  F.~Bougares, H.~Schwenk, and Y.~Bengio, ``On using monolingual corpora in
  neural machine translation,'' \emph{arXiv e-prints}, vol. abs/1503.03535,
  Mar. 2015.

\bibitem{watanabe2017language}
S.~Watanabe, T.~Hori, and J.~Hershey, ``Language independent end-to-end
  architecture for joint language identification and speech recognition,'' in
  \emph{IEEE Workshop on Automatic Speech Recognition and Understanding
  (ASRU)}, 2017, pp. 265--269.

\bibitem{kanda2016maximum}
N.~Kanda, X.~Lu, and H.~Kawai, ``Maximum a posteriori based decoding for {CTC}
  acoustic models,'' in \emph{Interspeech}, 2016, pp. 1868--1872.

\end{thebibliography}
\end{document}